\def\paperTitle{Data-Augmented Quantization-Aware Knowledge Distillation}

\def\authorBlock{
    Justin Kur \qquad
    Kaiqi Zhao \\
    Oakland University \\
    {\tt\small \{jakur, kaiqizhao\}@oakland.edu}
}

\newif\ifreview 
\newif\ifarxiv \newcommand{\arxiv}{\arxivtrue}
\newif\ifcamera 
\newif\ifrebuttal 

\arxiv

\pdfoutput=1
\documentclass[10pt,twocolumn,letterpaper]{article}
\ifreview \usepackage[review]{cvpr} \fi
\ifarxiv \usepackage[pagenumbers]{cvpr} \fi
\ifrebuttal \usepackage[rebuttal]{cvpr} \fi
\ifcamera \usepackage{cvpr} \fi

\usepackage{graphicx}	
\usepackage{amsmath}	
\usepackage{amssymb}	
\usepackage{booktabs}
\usepackage{times}
\usepackage{microtype}
\usepackage{epsfig}
\usepackage{caption}
\usepackage{float}
\usepackage{placeins}
\usepackage{color, colortbl}
\usepackage{stfloats}
\usepackage{enumitem}
\usepackage{tabularx}
\usepackage{xstring}
\usepackage{multirow}
\usepackage{xspace}
\usepackage{url}
\usepackage{subcaption}
\usepackage{xcolor}
\usepackage[hang,flushmargin]{footmisc}

\ifcamera \usepackage[accsupp]{axessibility} \fi

\newcommand{\nbf}[1]{{\noindent \textbf{#1.}}}

\usepackage{xr-hyper}

\makeatletter
\newcommand*{\addFileDependency}[1]{
  \typeout{(#1)}
  \@addtofilelist{#1}
  \IfFileExists{#1}{}{\typeout{No file #1.}}
}

\makeatother
\newcommand*{\myexternaldocument}[1]{
    \externaldocument{#1}
    \addFileDependency{#1.tex}
    \addFileDependency{#1.aux}
}

\definecolor{cvprblue}{rgb}{0.21,0.49,0.74}
\usepackage[pagebackref,breaklinks,colorlinks,allcolors=cvprblue]{hyperref}
\usepackage[capitalize]{cleveref}
\crefname{section}{Sec.}{Secs.}
\crefname{table}{Table}{Tables}
\crefname{figure}{Fig.}{Figs.}

\ifarxiv \crefname{appendix}{App.}{Apps.}
\else \crefname{appendix}{Suppl.}{Suppls.} \fi

\unless\ifarxiv \myexternaldocument{_supplementary} \fi

\newcommand{\boldhdr}[1]{\noindent \textbf{#1.}}

\begin{document}
\title{\paperTitle}
\author{\authorBlock}
\maketitle

\begin{abstract}
Quantization-aware training (QAT) and Knowledge Distillation (KD) are combined to achieve competitive performance in creating low-bit deep learning models. Existing KD and QAT works focus on improving the accuracy of quantized models from the network output perspective by designing better KD loss functions or optimizing QAT’s forward and backward propagation. However, limited attention has been given to understanding the impact of input transformations, such as data augmentation (DA). The relationship between quantization-aware KD and DA remains unexplored. 
In this paper, we address the question: how to select a good DA in quantization-aware KD, especially for the models with low precisions? 
We propose a novel metric which evaluates DAs according to their capacity to maximize the Contextual Mutual Information--the information not directly related to an image's label--while also ensuring the predictions for each class are close to the ground truth labels on average. 
The proposed method automatically ranks and selects DAs, requiring minimal training overhead, and it is compatible with any KD or QAT algorithm. 
Extensive evaluations demonstrate that selecting DA strategies using our metric significantly improves state-of-the-art QAT and KD works across various model architectures and datasets.
\end{abstract}

\section{Introduction}
\label{sec:intro}

Deep neural networks have shown outstanding performance in many areas of computer vision \cite{7298594, hu2018relation}. However, these networks contain many millions of parameters and require massive amounts of computation resources, which limits their usability in resource-constrained systems and edge devices \cite{peng2019correlationcongruenceknowledgedistillation}. 
Quantization is one of the important model compression approaches to address this challenge by converting the full-precision model weights or activations to lower precision.
In particular, Quantization-Aware Training (QAT) has achieved promising performance by simulating the effects of quantization during training and estimating the gradient through the quantized functions on the backward pass \cite{whitepaper}. 
Recent works \cite{sqakd, qkd, speq, ptg} show that the combination of QAT and knowledge distillation (KD) can improve the performance of the quantized models. However, the accuracy loss is often significant when the models are quantized to low precision \cite{qkd}. 

Most existing works on KD and QAT focus on improving the accuracy of quantized models by refining KD loss functions or optimizing QAT’s forward and backward propagation, with limited attention to the role of input transformations. 
Wang et al. \cite{wang2022makes} explored the interplay between KD and data augmentation (DA). 
However, their analysis primarily relies on a relatively simple metric that predicts the interplay between DA and model loss instead of accuracy, and it fails to effectively estimate the quantized model's accuracy (see \cref{sec:metric-val} for details). 
Moreover, their study evaluates individual DA methods in isolation, which is impractical for real-world applications where augmentations are typically used in combination. 
The relationship between quantization-aware KD and DA remains underexplored. 

In this work, we propose Data-Augmented Quantization-Aware Knowledge Distillation, an automated method for estimating the effectiveness of DA schemes for training low-precision models. Specifically, we construct a data augmentation ranking metric that maximizes Contextual Mutual Information (CMI), the information present in an image that is not directly related to its underlying class, while also minimizing the deviation between the average predictions for a class and its ground truth label.

The proposed metric has several advantages.
First, our metric can be computed cheaply, requiring only two passes through the training data, and no backward gradient computations or modifications to the full precision teacher model. This allows a large range of candidate DAs to be evaluated, and avoids a costly grid search as would often be performed when treating the augmentation choice as a hyperparameter. Second, because the metric is computed using only the full precision teacher model, it is inherently agnostic to the bit-width of the quantized student model. Third, the metric is capable of evaluating arbitrarily complex data augmentations. Because all that is needed to evaluate a DA is the full precision model's predictions and the ground truth image labels, any image transformation is admissible. 
For this work, we focus on well known DA policies, meaning we can quickly evaluate their transferability to new datasets in the context of QAT.

Our comprehensive evaluation shows that the proposed method improves the-state-of-art (SOTA) KD and QAT benchmarks under a variety of model architectures on CIFAR-10, CIFAR-100, and Tiny-ImageNet.
First, we evaluate 7 well-known DAs and find that the chosen DA with our metric improves top-1 test accuracy of existing SOTA QAT methods, e.g., EWGS, PACT, LSQ, and DoReFa, by up to 9\% under diverse quantization levels.
Second, our DA improves Top-1 test accuracy across a variety of KD algorithms, improving performance on some algorithms like CRD \cite{crd} by up to 3\% with identical training hyperparameters. 
Third, we achieve superior Top-1 test accuracy on multiple datasets compared to existing approaches that integrate KD with QAT, including QKD \cite{qkd}, SPEQ \cite{speq}, and SQAKD \cite{sqakd}.

Our main contributions can be summarized as: 1) We introduce data augmentation search into the training loop of knowledge distillation assisted QAT. 2) We construct a metric that outperforms existing heuristics at predicting the effectiveness of a data augmentation in quantizing a model using KD. Importantly, this heuristic is cheap to compute, so it can be effectively used to reduce the search space of DAs.

\section{Related Work}
\label{sec:related} 
\nbf{Knowledge Distillation (KD)}
KD transfers knowledge from large networks (termed ``teacher'') to improve the performance of small networks (termed ``student'').
KD was originally introduced by Hinton et al. \cite{kd}, as an efficient method for encoding the knowledge of an ensemble of models into a single model by using the logits of the ensemble teacher as the target for the student. Since then, many variations on the original idea have been proposed. CRD \cite{crd} and RKD \cite{rkd} instead attempt to match the structure of the representations at the last layer of the teacher and student. CRD achieves this by means of contrastive learning and RKD by directly using the n-tuple angle and distances between data samples. Other methods try to match the representations at the inner layers of the networks such as Attention Transfer \cite{at} and Neuron Selectivity Transfer \cite{nst}. In contrast to these approaches, we do not seek to modify the underlying algorithms of KD, but rather integrate KD into model quantization, so our method is orthogonal to these works. 

\nbf{Quantization Aware Training}
Network quantization can be divided into two categories: post training quantization (PTQ) and QAT. PTQ methods do not retrain or fine tune the model, instead quantizing the weights directly to minimize the degradation from the loss in precision. This can be done using training data, or in a data-free manner \cite{gholami2022survey}.  In contrast, in QAT the quantization is performed alongside the training of the network. QAT is more expensive than post training quantization, but in return it generally incurs a smaller penalty to model accuracy \cite{gholami2022survey}. In this work, we focus on improving QAT methods through the addition of KD. 

Zhou et al. \cite{dorefa} introduce DoReFa-Net which expands upon initial research on binary neural networks \cite{courbariaux2016binarizedneuralnetworkstraining} \cite{rastegari2016xnornetimagenetclassificationusing}, learning a constant scalar to scale all quantized convolution weights, and using straight-through estimation on activations. Later works introduce more learnable parameters for quantization, including activation clipping parameters in PACT \cite{pact} and LSQ \cite{lsq}, and parameters that control the interval between quantization levels in EWGS \cite{ewgs} and QIL \cite{jung2018learningquantizedeepnetworks}. These QAT methods have been shown to reduce the accuracy loss from quantization, but the gap between the quantized and the full precision models can still be large, especially when models are quantized to less than 8 bits. Our work aims to mitigate the accuracy loss of existing quantization algorithms through self-supervised knowledge distillation. 

\nbf{Data Augmentation Search}
Data augmentations are widely used when training deep neural networks on vision tasks, but automatically determining the best data augmentation for a given dataset or model is challenging and often computationally intensive. AutoAugment \cite{autoaugment} uses reinforcement learning to find a set of augmentations that performed well on CIFAR, ImageNet, and SVHN. Fast AutoAugment \cite{lim2019fastautoaugment} avoids the full training process and instead performs density matching between the augmented and unaugmented images. RandAugment \cite{rand} reduces the search space of augmentations by applying any augmentation from the set with uniform probability, and learns a single strength value that applies to all data augmentations. TrivialAugment \cite{trivial} removes the search parameters altogether, instead opting to uniformly sample an augmentation from its set, and uniformly sample the strength with which the augmentation should be applied. Nonetheless, it can improve performance on standard models and datasets over augmentations with thousands of GPU-hours of search. AugMix \cite{augmix} searches for augmentations to try to overcome dataset drift between the training and test sets. Other papers build upon the original scope of augmentation search, such as Population Based Augmentation \cite{ho2019population}, a method for learning a curriculum of data augmentations rather than a single augmentation that lasts for the entirety of training. Adversarial data augmentation search \cite{zhang2019adversarialautoaugment} \cite{9880344} incorporates adversarial learning into data augmentation search to optimize a data augmentation policy. Unlike these works, we do not focus on building a new data augmentation policy from scratch, but we do look to prune the search space of existing augmentations \textit{without} training the student model. Furthermore, we specifically define our DA evaluation metric for the case of KD.

There is also a smaller body of work focusing on data augmentations for KD. Wang et al. \cite{good} provide a statistical justification for using the DA that minimizes the variance of the teacher's predictions. Accordingly, we benchmark our DA selection for quantization aware KD against this methodology, and find that our method is superior for models quantized to low bit precision, as shown in \cref{sec:metric-val}. 

\nbf{Knowledge Distillation in QAT}
Recent works have integrated KD into quantized model training, but the loss in model accuracy at low precision is still a significant limitation. QKD \cite{qkd} proposes a three phase curriculum where first the quantized model is trained without KD, then the teacher and student learn together, and finally normal KD between the student and teacher occurs.  SPEQ \cite{speq} uses a teacher whose activations are stochastically quantized to different bit levels, treating the teacher as an ensemble. CMT-KD \cite{cmtkd} optimizes several quantized teachers and students collaboratively, performing distillation at both the logit level and in intermediate network layers. PTG \cite{ptg} progressively quantizes its models into lower bit precisions, and concurrently trains the teacher alongside the student. Zhao et al. \cite{sqakd} propose Self-Supervised Quantization Aware Knowledge Distillation (SQAKD) which initializes a quantized student model to its teachers weights, and perform self-supervised KD without using loss from labels. These approaches have combined KD and QAT, but they have not considered the effects of DA at distillation time, nor integrating DA search into the quantization pipeline.

\section{Methodology}
\label{sec:method}
\subsection{Preliminaries}
Importantly, in KD the model with the highest training or validation accuracy may not serve as the best teacher \cite{cho2019efficacy}, which motivates the development of heuristics specific to knowledge distillation. Ye et al. introduce the concept of Contextual Mutual Information (CMI) \cite{cmi}, which provides a quantifiable metric for the dark knowledge that the teacher can impart upon the student. Informally, an image can thought of as containing two parts--firstly is the part that is essential to its class label, and secondly is the background context of the image. Following existing terminology \cite{cmi}, we call the essential class information the prototype of the image, and the second part the contextual mutual information. 
In other words, the CMI is the information obtained about an image using the teacher model's prediction, given that the ground truth is already known. This corresponds precisely with the information in the image that is not directly tied to its ground truth label. Intuitively, if a teacher model could provide a large CMI target for its student, then it would offer value beyond what was already present in the ground truth labels.  
Ye et al. \cite{cmi}, provide the following empirical method for computing the CMI of a teacher model: 
\begin{equation}
CMI_{emp}(f) = \frac{1}{N} \sum_{y \in [C]} \sum_{x_j \in D_y} KL(P_{x_{j}} \parallel Q^{y}_{emp})
\end{equation}
\begin{equation}
\text{where } Q^{y}_{emp} = \frac{1}{\vert D_y \vert} \sum_{x_j \in D_y} P_{x_{j}} \text{, for } y \in [C]
\end{equation}

Rather than directly fine tune a teacher with CMI as a loss term as in previous works, we instead search for a DA that will modify the training data to induce the desired balance of CMI and log likelihood at distillation time. The rest of this section focuses on how we concretely achieve that goal. 

\subsection{Quantitatively Evaluating DAs for Distillation}
\label{sec:metric}

We observe that rather than fine-tuning a model with CMI incorporated into the loss term, it is also sensible to hold the model fixed, and modify the data to induce a better balance of log likelihood and CMI. Concretely, this can be achieved by applying data augmentations to the underlying dataset $D$ at distillation time. In this way, we eliminate the need for an additional model training procedure, and we avoid the decline in teacher validation accuracy \cite{cmi} that results from incorporating a secondary term into the loss function. 

However, any metric which incorporates CMI must also include a factor that rewards the teacher model's faithfulness to the ground truth labels, because CMI alone only considers the \textit{diversity} and not the \textit{accuracy} of predictions. Informally, the CMI indicates that the teacher is providing information for each image that is not present in the underlying class prototype, but the metric should also consider that the prototype for any given class should not deviate too much from its ground truth value (a one hot encoding of the label). The average prediction for a given class $y$ is exactly the term $Q^{y}_{emp}$, so we can quantify the difference in the following way using their \textbf{centroid deviation} (DEV): 
\begin{equation}
DEV(f) = \frac{1}{C}\sum_{y \in [C]}KL(1_y \parallel Q^{y}_{emp})
\end{equation}
where $1_y$ indicates the one-hot encoded vector corresponding to class y. Our metric should aim to maximize the information gain from CMI, while minimizing the relative entropy between the class prototypes and the ground truth prototypes. Thus, the two terms can be combined as follows by minimizing the following metric $M(f)$:
\begin{equation}
M(f) = DEV(f) - CMI_{emp}(f)
\end{equation}
Both CMI and DEV are calculated as averages of KL divergences, so $M(f) \in (-\infty, \infty)$, but a model which exactly predicts its labels would only achieve a score of 0, rightly indicating that it would provide no additional value to distillation when ground truth labels are available during training. 

\subsection{Handling Multiple Labels per Image}
Although the above formulation works in the usual case of single-label multiclass classification, it is necessary to generalize the above definition of $CMI_{emp}$ if we want to accommodate certain classes of augmentations such as CutMix \cite{cutmix} and MixUp \cite{mixup}, that combine multiple images into one. For instance, in CutMix, two images A and B are combined together and the resulting ground truth of the training example is set to $y = \lambda y_A + (1 - \lambda)y_b$ \cite{cutmix}, with $\lambda$ determined by the proportion of the training image originating from image A. When images A and B correspond to different classes, their $CMI_{emp}$ would be undefined. As such, we introduce the following generalization of $CMI$ that we call generalized CMI or $GCMI$: 
\begin{equation}
GCMI_{emp}(f) = \frac{1}{N}\sum_{x_i \in D} KL(P_{x_{i}} \parallel Q^{i}_{emp})
\end{equation}
\begin{equation}
\text{where } Q^{i}_{emp} = \sum_{j \in [C]} P_{x_{ij}} \cdot \frac{Z_j}{||Z_j||}, Z_j = \sum_{x_k \in D} w_{j_k} P_{x_{k}}
\end{equation}
where the new term $w_{j_k}$ indicates the label weight corresponding to class $j$ for image $k$ under the given augmentation. This generalization would work for arbitrary target probability vectors, but as an example in the CutMix case, if an image $i$ was $70\%$ $c_1$ and $30\%$ $c_2$ then its $Q^{i}_{emp}$ would be constructed from $0.7 Z_{c_{1}} + 0.3 Z_{c_{2}}$ after the proper normalization of the $Z$ terms. Likewise, during the computation of the $Z$ values, the image would contribute $0.7$ times its value to the $c_1$ prototype and $0.3$ times its value to the $c_2$ prototype. 

Returning to the metrics from \ref{sec:metric}, $M(f)$ can be calculated by replacing the CMI term with the new generalized $GCMI_{emp}$. In the calculation of $DEV$, we can replace the $Q^y_{emp}$ term in with $\frac{Z_j}{||Z_j||}$, which now corresponds to the empirically calculated prototype for each class. This allows our metric to evaluate the full category of batch-level DAs including MixUp and CutMix. 

\nbf{Practical Usage}
Once a full precision model is trained (or otherwise obtained), it will be considered the teacher for the quantized model. Before knowledge distillation is performed, the above metric $M(f)$ should be computed for each candidate DA. This requires the average prototype for each class $Q^{y}_{emp}$, which can be computed in one forward pass through the training data. Then the CMI for each sample can be computed with one more pass through the training data. In most cases, two passes through the dataset is cheap enough that the candidate DAs can be exhaustively searched, but for larger datasets a randomly sampled subset will also suffice to approximate the prototypes and average CMI values. When the best DA according to our metric is selected, then self-supervised KD and QAT can begin.

\section{Evaluation}
\label{sec:evaluation}

\boldhdr{Datasets and models}
We evaluate our methodology across a range of datasets and model architectures. CIFAR-10 \cite{cifar} is a standard computer vision benchmark consisting of 50,000 training images and 10,000 test images of size $32 \times 32$ with 10 total classes. On CIFAR-10, we evaluate our performance on VGG-8 and ResNet-20. CIFAR-100 \cite{cifar} has the same number of images and image characteristics as CIFAR-10, but with 100 total classes. We train deeper models as compared to CIFAR-10, using VGG-13 and ResNet-32 as representative members of these two architectures. Finally, Tiny ImageNet \cite{tinyimagenet} consists of a subset of 100,000 training images from the ImageNet Large Scale Visual Recognition Challenge \cite{imagenet}. It contains 200 classes, and downsamples the original images to $64 \times 64$. Tiny ImageNet presents a more computationally challenging task than CIFAR, while remaining smaller than the full ImageNet-1K dataset. We utilize both the MobileNetV2 and ResNet-18 architectures to benchmark our performance on Tiny ImageNet.

\boldhdr{Baselines}
We compare our results against established baselines in the following four categories:
\begin{itemize}
\item \textbf{QAT}: PACT \cite{pact}, DoReFa \cite{dorefa}, LSQ \cite{lsq}, EWGS \cite{ewgs}. 
\item \textbf{KD}: AT \cite{at}, CC \cite{cc}, CRD \cite{crd}, NST \cite{nst}, RKD \cite{rkd}, SP \cite{sp}. 
\item \textbf{KD + QAT}: QKD \cite{qkd}, SPEQ \cite{speq}, SQAKD \cite{sqakd}. 
\item \textbf{DA Search in KD}: Minimizing the variance of the teacher's predictions \cite{good}. 
\end{itemize}
For QAT, we compare our accuracy results to those reported from using QAT alone. For KD, we consider the quantized accuracy resulting from our selected DA against the accuracy using limited DAs. For KD + QAT, we compare against cited results with identical model architectures and datasets. 
Finally, for DA search we consider the quantized accuracy for the highest scoring DA by each metric. 

\boldhdr{Data augmentation schemes}
We focus primarily on data augmentation techniques which are automatic, which is to say they were found through a data-driven algorithm rather than hand-crafted. Specifically, we investigate the following standard data augmentation policies: 
\begin{itemize}
    \item \textbf{Auto Augment (AA)} \cite{autoaugment}: A landmark work in data augmentation search, this method used reinforcement learning to develop augmentation policies for datasets including CIFAR and ImageNet. We consider the CIFAR and ImageNet policies separately in this paper, as they are quite different even though they resulted from the same search algorithm. 
    \item \textbf{RandAugment (Rand)} \cite{rand}: An automatic augmentation method found using much less computation than Auto Augment. 
    \item \textbf{Trivial Augment (TA) \cite{trivial} } An automatic augmentation method that uses no search at all--instead, it randomly samples an augmentation from its set and randomly samples its strength for each image individually. 
    \item \textbf{AugMix} \cite{augmix}: An automatic augmentation optimized for robustness under data distribution shift. 
    \item \textbf{CutMix \cite{cutmix}}: A batch-level augmentation method that combines two images from a batch by copy-pasting one rectangular area from one image into another image. The label of the resulting image is set by interpolating between the labels of the two originals according to their relative sizes. 
    \item \textbf{Minimal}: A conservative augmentation method including random crop, random flip, and color jitter for Tiny ImageNet only. All our methods use this minimal augmentation as a preprocessing step, but we also include it as a standalone DA. 
\end{itemize}

\boldhdr{Implementation details}
Experiments are carried out on Python 3.11 using PyTorch 2.1 running on RTX 3000 series GPUs. Full precision models on CIFAR are trained using the minimal augmentation described above and CutMix. Full precision models on Tiny ImageNet are trained using only the minimal augmentation. 

We use the common notation ``$W{*}A\times$'' to denote that the weights and activations are quantized into $*$bit and $\times$bit, respectively. For our experiments, we quantize weights and activations to the same bit widths, but in general the two precisions could differ, so we include both values for clarity. 

All CIFAR experiments use the EWGS \cite{ewgs} quantizer with the weight hyperparameters for labels and KD set to 1.0 and 2.0 respectively. For CIFAR-10, all models are trained with the ADAM optimizer with a learning rate of 1e-3, with a cosine learning rate decay, and a batch size of 256. EWGS has its own separate optimizer, but it is initialized in the same way except the initial learning rate is set to 1e-5. The model's weight decay is set to 1e-4. For VGG-8, 400 epochs of quantization and distillation are performed. For ResNet-20, 1200 epochs are performed. For CIFAR-100 all learning rates are halved, the batch size is 64, and all models are trained for 200 epochs.

On Tiny ImageNet, we quantize models using DoReFa \cite{dorefa}, PACT \cite{pact}, and LSQ \cite{lsq} with implementations provided by MQBench \cite{mqbench}. Models are optimized using SGD with the initial learning rate set to 4e-3 with 2500 warmup steps and cosine learning rate decay. Momentum is set to 0.9 and weight decay is set to 1e-4. The batch size is 64, and all models are trained for 100 epochs. During training, all images are upscaled from their original $64 \times 64$ to $224 \times 224$ before augmentations are applied. Validation images are upscaled to $256 \times 256$. For ResNet-18, we keep the same loss term hyperparameters found on CIFAR, setting the label loss weight to 1.0 and the KD loss weight to 2.0. On MobileNetV2, we keep the same ratio, but increase the label loss weight to 3.0 and the KD loss weight to 6.0.

\subsection{Empirical validation of the proposition} \label{sec:metric-val}

\begin{figure}

\centering
\subfloat[CIFAR-10 ResNet-20 W2A2]{\label{fig:aug-cifar10}
\centering
\includegraphics[width=0.9\linewidth]{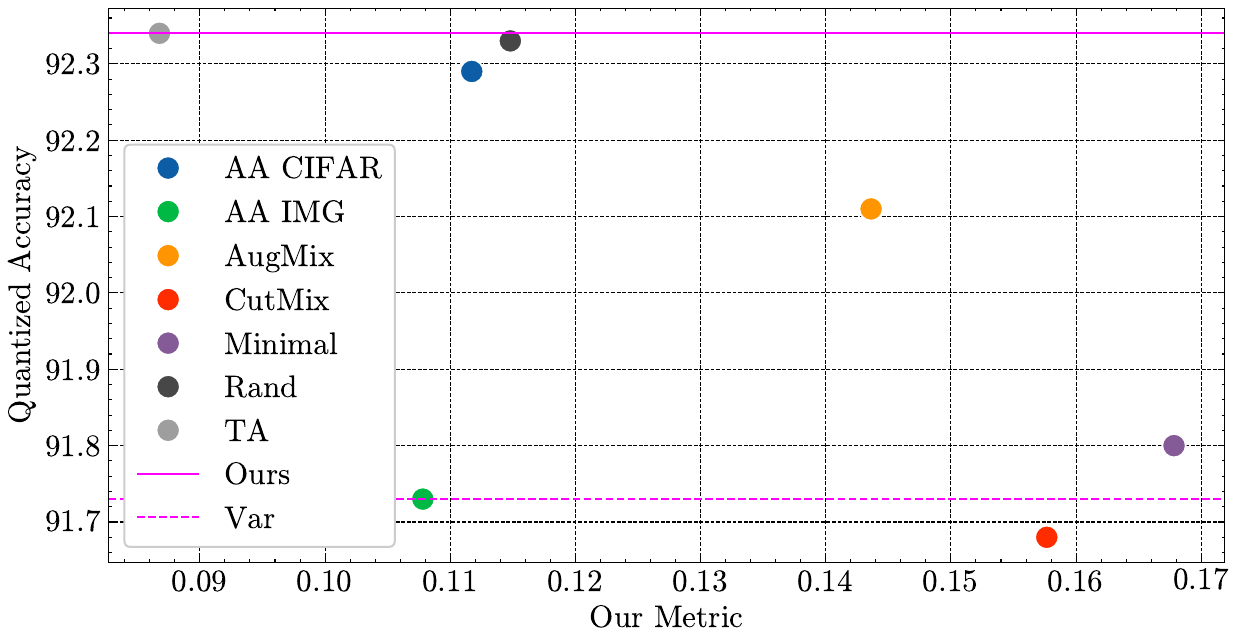}
}
\hfill
\subfloat[CIFAR-100 VGG-13 W3A3]{\label{fig:aug-cifar100}
\centering

\includegraphics[width=0.9\linewidth]{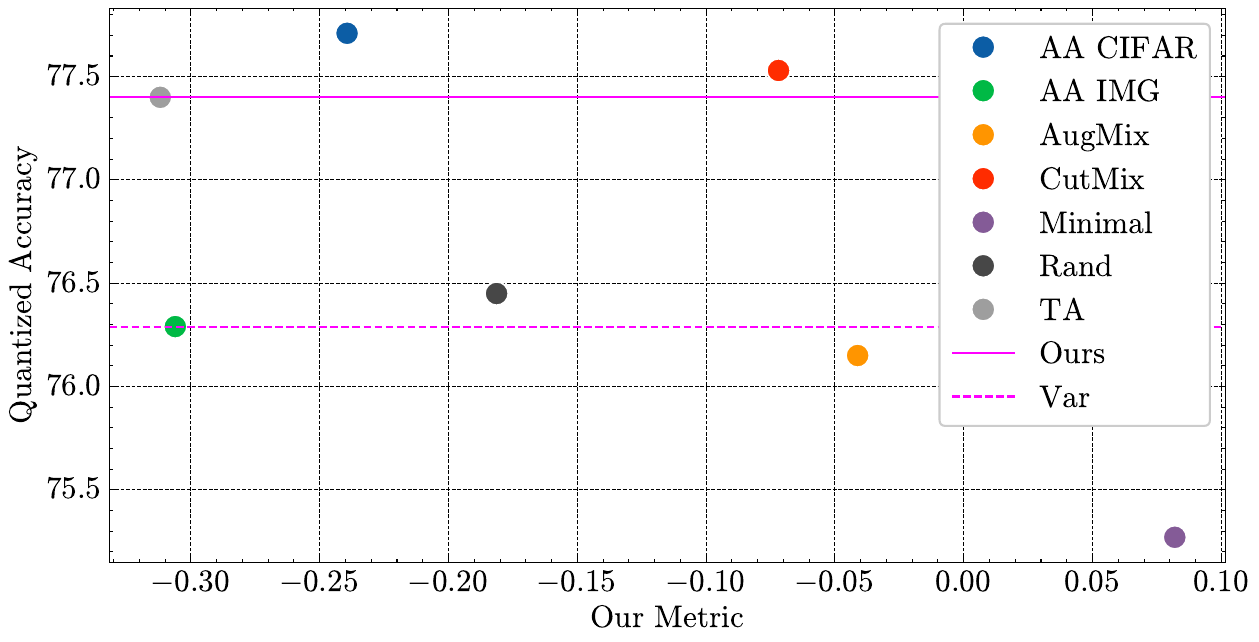}
}
\hfill
\subfloat[Tiny ImageNet MobileNetV2 W3A3]{\label{fig:aug-tiny}
\centering
\includegraphics[width=0.9\linewidth]{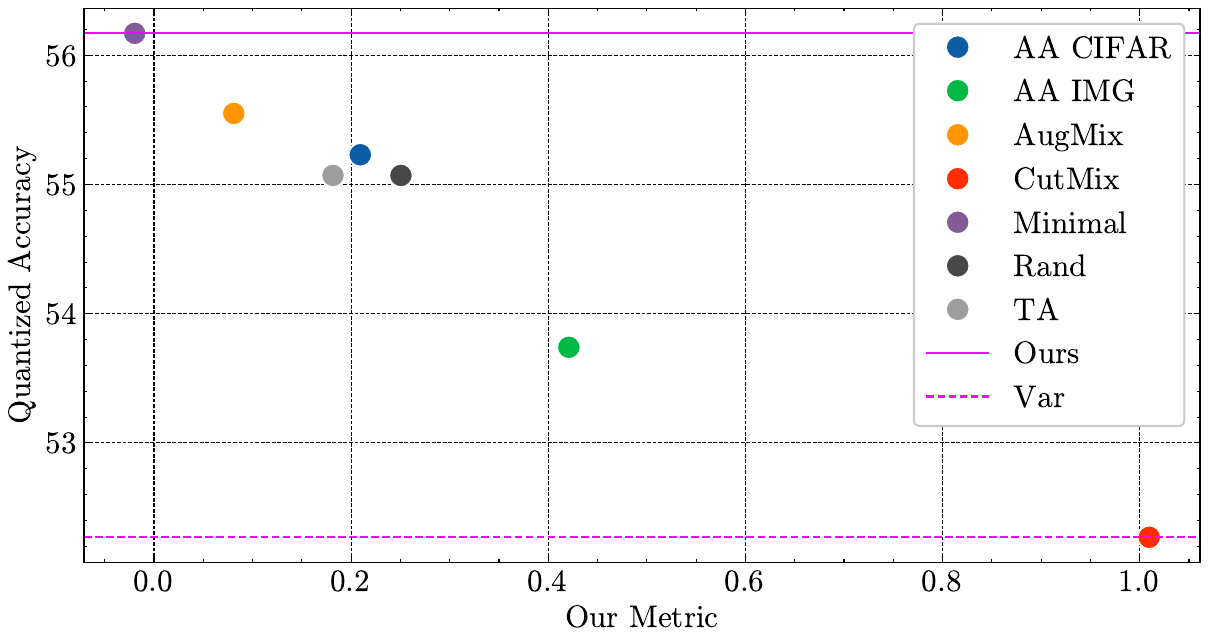}
}
\hfill

\caption{Scatter visualization of the quantized model's top-1 test accuracy vs. our metric for 7  data augmentation (DA) schemes on CIFAR-10, CIFAR-100 and Tiny-ImageNet. The X-axis represents the DA's score by our metric (lower is better). The Y-axis indicates the quantized model's Top-1 test accuracy. The accuracy resulting from our approach and the baseline minimal variance method \cite{good} are marked with horizontal lines. }
\label{fig:aug_search}

\end{figure}

\cref{fig:aug_search} outlines our augmentation search results across 3 different models and datasets, showing that our DA selection outperforms the existing baseline of selecting DA based on the variance observed on training data \cite{good}. \cref{fig:aug-cifar10} shows our results on CIFAR-10 using the trained full precision ResNet-20 model to collect the necessary CMI, centroid deviation, and variance statistics. Our metric correctly predicts Trivial Augment will be the best augmentation, achieving 92.34\% Top-1 accuracy. The variance metric meanwhile selects Auto Augmentation ImageNet policy, which at 91.73\% underperforms even the approach with minimal augmentations. 

\cref{fig:aug-cifar100} demonstrates the performance on CIFAR-100 using the VGG-13 architecture. Although our metric does not select the absolute best method in ranking, the difference between our selected method (Trivial Augment) and the best method is only 0.3\%; meanwhile, the Auto Augmentation ImageNet augmentation selected by variance is a full 1.4\% lower than the best method. We note that although the statistics were recomputed for CIFAR-10 and CIFAR-100, the overall ranking for both datasets is similar. Empirically, we observe that our ranking results closely match across model architectures, but not across datasets. However, considering the similarity of CIFAR-10 and CIFAR-100 datasets, the rankings overlapping here is not entirely unexpected. 

\cref{fig:aug-tiny} provides our DA search results on Tiny ImageNet using MobileNetV2. On Tiny ImageNet, our predictions substantially differ from the variance metric: our metric chooses the best performing DA while checking for variance chooses the worst. In this case, our metric chooses the simple minimal augmentation which employs cropping, horizontal flipping, and color jitter, and we find that this method outperforms the next best DA AugMix by more than 0.6\%. On the other hand CutMix, the method selected by inspecting the teacher's variance, performs very poorly, which is surprising given previous works have used it successfully in KD training \cite{good}. However, we can see that it was the second best method for CIFAR-100 on VGG-13, so perhaps it can still be useful in quantization aware distillation, given the right model and dataset. 

\cref{tab:spearman} reports the overall Spearman correlation for our DA predictions compared against the predictions from minimizing variance, where each individual sample represents a DA and its corresponding quantized student perfomance. Because both our metrics are ``lower is better'', the ideal Spearman value is $-1.0$, with $1.0$ being the worst possible score. Our method reports a better Spearman correlation on the examined architectures on CIFAR-10, CIFAR-100, and Tiny ImageNet, indicating that our average DA ranking is more accurate than the variance metric. In particular, we find that our methodology is superior for predicting the quantized distillation performance on MobileNetV2 on Tiny ImageNet, where our method gives a nearly perfect ranking. On the other hand, using variance gives roughly the reverse of the actual ranking, showing its poor performance in this scenario. 

\begin{table}[htb]
\centering
\caption{Spearman correlation for each DA's predicted vs. actual performance (lower is better). We benchmark our methodology against the variance metric (var). $-1$ indicates perfect negative correlation, which is the goal of both methods. }
\label{tab:spearman}
\begin{tabular}{@{}llll@{}}
\toprule
 & CIFAR-10  & CIFAR-100 & Tiny ImageNet \\ \midrule
 & ResNet-20 & VGG-13    & MobileNetV2   \\ \midrule
Var \cite{good}& \multicolumn{1}{r}{-0.429}          & \multicolumn{1}{r}{-0.464}          & \multicolumn{1}{r}{0.883}           \\
Ours     & \multicolumn{1}{r}{\textbf{-0.536}} & \multicolumn{1}{r}{\textbf{-0.536}} & \multicolumn{1}{r}{\textbf{-0.937}} \\ \bottomrule
\end{tabular}
\end{table}

\subsection{Improving SOTA QAT}
\boldhdr{Results on CIFAR-10 and CIFAR-100}
Using Trivial Augment, the DA which is selected in \cref{sec:metric-val}, we perform model quantization on CIFAR-10 and CIFAR-100. The results are presented in \cref{tab:cifar_quant}. In all cases, our method outperforms the existing baselines presented by direct QAT with EWGS, and KD-assisted QAT. Our results are most pronounced at the 4 bit quantization level, with improvements of over 3\% compared to existing methods on CIFAR-100 for both VGG-13 and ResNet-32. Additionally, at the 4 bit quantization level our student model achieves higher accuracy than its full precision teacher on all CIFAR experiments, which is not reported for either EWGS or SQAKD \cite{sqakd}. Although improvements are smaller at the 2 bit quantization level, the difference between our results and existing work is still significant, including over 1\% improvement on CIFAR-10 VGG-8 and nearly 2\% improvement on CIFAR-100 VGG-13. These findings validate the reliability of our DA selection for the CIFAR datasets.

\begin{table*}[htb]
\centering
\caption{Top-1 test accuracy (\%) on CIFAR-10 and CIFAR-100. ``$W{*}A\times$'' denotes that the weights and activations are quantized into $*$bit and $\times$bit, respectively. Full-precision models' accuracies are as follows: VGG-8 (92.91), ResNet-20 (94.00), VGG-13 (76.24), ResNet-32 (74.57). The results of the compared methods including EWGS~\cite{ewgs} and SQAKD~\cite{sqakd} are cited from their original papers.}
\label{tab:cifar_quant}
\begin{tabular}{@{}llrrrlrrr@{}}
\toprule
\multicolumn{1}{|l|}{Dataset} &
  \multicolumn{4}{c|}{CIFAR-10} &
  \multicolumn{4}{c|}{CIFAR-100} \\ \midrule
\multicolumn{1}{|l|}{Model} &
  \multicolumn{2}{c|}{VGG-8} &
  \multicolumn{2}{c|}{ResNet-20} &
  \multicolumn{2}{c|}{VGG-13} &
  \multicolumn{2}{c|}{ResNet-32} \\ \midrule
\multicolumn{1}{l|}{Quantization} &
  \multicolumn{1}{l|}{W2A2} &
  \multicolumn{1}{l|}{W4A4} &
  \multicolumn{1}{l|}{W2A2} &
  \multicolumn{1}{l|}{W4A4} &
  \multicolumn{1}{l|}{W2A2} &
  \multicolumn{1}{l|}{W4A4} &
  \multicolumn{1}{l|}{W2A2} &
  \multicolumn{1}{l}{W4A4} \\
\multicolumn{1}{l|}{EWGS} &
  \multicolumn{1}{r|}{90.84} &
  \multicolumn{1}{r|}{90.95} &
  \multicolumn{1}{r|}{91.41} &
  \multicolumn{1}{r|}{92.40} &
  \multicolumn{1}{r|}{73.31} &
  \multicolumn{1}{r|}{73.41} &
  \multicolumn{1}{r|}{69.37} &
  \multicolumn{1}{r}{70.50} \\
\multicolumn{1}{l|}{SQAKD (EWGS)} &
  \multicolumn{1}{r|}{91.55} &
  \multicolumn{1}{r|}{91.31} &
  \multicolumn{1}{r|}{91.80} &
  \multicolumn{1}{r|}{92.59} &
  \multicolumn{1}{r|}{74.65} &
  \multicolumn{1}{r|}{74.67} &
  \multicolumn{1}{r|}{69.99} &
  \multicolumn{1}{r}{71.65} \\
\multicolumn{1}{l|}{Ours (EWGS)} &
  \multicolumn{1}{r}{\textbf{92.66}} &
  \textbf{93.66} &
  \textbf{92.34} &
  \textbf{94.30} &
  \multicolumn{1}{r}{\textbf{76.59}} &
  \textbf{77.97} &
  \textbf{70.20} &
  \multicolumn{1}{r}{\textbf{75.11}} \\ \bottomrule
\end{tabular}
\end{table*}

\boldhdr{Results on Tiny ImageNet}
For Tiny ImageNet, our DA selection algorithm selects only the Minimal augmentation method, as can be seen in \cref{fig:aug-tiny}. In spite of the simplicity of the augmentation, however, our method improves quantized accuracy in all tested cases, as reported in \cref{tab:tiny_quant}. In the most extreme case of ResNet-18 PACT W4A4, our method improves accuracy over SQAKD by almost 6\%. As in CIFAR, all over our models quantized to W4A4 or greater achieve superior performance than their full precision teachers. Meanwhile, none of the SQAKD student models listed here achieve this feat, as their teacher Top-1 accuracies on MobileNetV2 and ResNet-18 are 58.07\% and 65.59\% \cite{sqakd}. Because our method selects relatively limited DAs for Tiny ImageNet, the biggest difference between our training and SQAKD's is our use of cross entropy loss from labels as a component of the loss function, which appears beneficial to model performance. These experiments on Tiny ImageNet also show that our method is robust to multiple quantizers, as the top-1 accuracy significantly improves over existing results under PACT, LSQ, and DoReFa. 

\begin{table}[htb]
\caption{Top-1 test accuracy (\%) on Tiny ImageNet using MobileNetV2 and ResNet-18. QAT refers to quantization using QAT only, i.e. by PACT \cite{pact}, LSQ \cite{lsq}, or DoReFa \cite{dorefa}. Both results are as reported in the SQAKD paper \cite{sqakd}. Our full-precision accuracies are: MobileNetV2 (58.64) and ResNet-18 (66.87)}
\label{tab:tiny_quant}
\begin{tabular}{@{}lrrr@{}}
\toprule
                      & \multicolumn{1}{l}{Ours} & \multicolumn{1}{l}{SQAKD} & \multicolumn{1}{l}{QAT} \\ \midrule
MobileNetV2 PACT W3A3 & \textbf{56.17}           & 52.73                     & \multicolumn{1}{r}{47.77}                     \\
MobileNetV2 PACT W4A4 & \textbf{59.48}           & 57.14                     & \multicolumn{1}{r}{50.33}                     \\
ResNet-18 PACT W3A3   & \textbf{62.31}           & 61.34                     & \multicolumn{1}{r}{58.09}                     \\
ResNet-18 LSQ W3A3    & \textbf{66.47}           & 65.21                     & \multicolumn{1}{r}{61.99}                     \\
ResNet-18 PACT W4A4   & \textbf{67.39}           & 61.47                     & \multicolumn{1}{r}{61.06}                     \\
ResNet-18 DoReFa W8A8 & \textbf{67.19}           & 64.88                     & \multicolumn{1}{r}{63.23}                     \\ \bottomrule
\end{tabular}
\end{table}

\subsection{Improving SOTA KD}
\label{sec:kd}
Although we have focused on using standard KD thus far, \cref{tab:kd} demonstrates our method's efficacy on other competitive KD algorithms. Our selected DA Trivial Augment successfully outperforms the baseline of no augmentations for all 7 evaluated methods. Each experiment is run for 200 epochs, with the loss weight hyperparameters set according to what is proposed in their original papers. Every method includes loss from ground truth labels, except Pure KD, which we specifically use to mean KD with no additional loss terms. Our DA selection improves the accuracy of CRD by over 3\%, as well as improving the accuracy on NST, RKD, and SP by over 1\%. Interestingly, the smallest improvement is on pure KD, indicating that heavy DA training benefits more from having label information present. Intuitively this makes sense, because effective DAs for normal model training need to preserve the original class of the transformed image. 

\begin{table}[htb]
\centering
\caption{Top-1 test accuracy comparing our selected DA against limited baseline DA across different methods of knowledge distillation using CIFAR-100 VGG-13 at W2A2. Full precision accuracy: 76.24\%.}
\label{tab:kd}
\begin{tabular}{@{}lrr@{}}
\toprule
        & \multicolumn{1}{l}{Limited DA} & \multicolumn{1}{l}{Ours} \\ \midrule
AT \cite{at}     & 75.49                        & \textbf{75.84}           \\
CC  \cite{cc}   & 73.51                        & \textbf{73.96}           \\
CRD  \cite{crd}   & 73.96                        & \textbf{77.16}           \\
NST  \cite{nst}   & 74.85                        & \textbf{76.33}           \\
RKD \cite{rkd}    & 74.05                        & \textbf{75.25}           \\
SP \cite{sp}     & 74.61                        & \textbf{76.29}           \\
Traditional KD & 72.45                        & \textbf{72.66}           \\ \bottomrule
\end{tabular}
\end{table}

\subsection{Comparison with SOTA Methods Applying both QAT and KD}
\cref{tab:meta} demonstrates our performance compared to several preexisting works integrating KD with QAT: namely QKD \cite{qkd}, SQAKD \cite{sqakd}, and SPEQ \cite{speq}. In every case we outperform the Top-1 accuracies reported in these papers. Our results use Trivial Augment, the DA selected by our metric on both CIFAR-10 and CIFAR-100. Notably, we exceed the performance of QKD by over 1\% on CIFAR-10 and almost 2\% on CIFAR-100 at the W4A4 quantization level. Although SPEQ does not report its results at the W3A3 or W4A4 levels, we can see that we outperform its benchmarks by roughly 1\% on both CIFAR-10 and CIFAR-100 at the W2A2 quantization level. This demonstrates that our method performs well across a range of quantization levels. 

\begin{table}[htb]
\caption{Top-1 test accuracy comparison on CIFAR between our method and several works integrating KD with QAT. All results are cited from their original papers. }
\label{tab:meta}
\begin{tabular}{@{}lrr|rrr@{}}
\toprule
\multicolumn{1}{l|}{}      & \multicolumn{2}{c|}{CIFAR-10}   & \multicolumn{3}{c|}{CIFAR-100}                   \\ \cmidrule(l){2-6} 
\multicolumn{1}{l|}{Model} & \multicolumn{2}{c|}{ResNet-20}  & \multicolumn{3}{c|}{ResNet-32}                   \\ \cmidrule(l){2-6} 
Quant & \multicolumn{1}{l}{W2A2} & \multicolumn{1}{l}{W4A4} & \multicolumn{1}{l}{W2A2} & \multicolumn{1}{l}{W3A3} & \multicolumn{1}{l}{W4A4} \\
QKD                        & 90.5           & 93.1           & 66.4           & 72.2           & \multicolumn{1}{r}{73.3}           \\
SPEQ                      & 91.4           & -              & 69.1           & -              & \multicolumn{1}{r}{-}              \\
SQAKD                      & 91.8           & 92.59          & 69.99          & -              & \multicolumn{1}{r}{71.65}          \\
Ours                       & \textbf{92.34} & \textbf{94.30} & \textbf{70.20} & \textbf{74.09} & \multicolumn{1}{r}{\textbf{75.11}} \\ \bottomrule
\end{tabular}
\end{table}

\section{Ablation Study}

\boldhdr{Effects of Augmentations}
\cref{tab:cifar-ablate} compares our methodology to models trained under the same conditions but with limited DAs. All ablation study results are calculated using 200 epochs of QAT training. On CIFAR-10 and CIFAR-100, we benchmark against a DA which only performs random horizontal flips and image cropping with padding of 4 pixels. Because CIFAR classes are invariant to horizontal flips, this is a very safe and common DA, and the random cropping only adds a small amount of noise to the image by translating it slightly. This data augmentation is used in other KD under QAT methodologies \cite{sqakd}, so it makes a good baseline comparison. As we can see, our selected DA outperforms the baseline in all cases except one. In some cases, it improves accuracy by upwards of 2\%, despite using the same teacher model. 

\begin{table}[htb]
\centering
\caption{Top-1 accuracy comparing our chosen Trivial Augment against a very limited DA for CIFAR-10 and CIFAR-100. }
\label{tab:cifar-ablate}
\begin{tabular}{@{}lrr@{}}
\toprule
                              & \multicolumn{1}{l}{Limited DA} & \multicolumn{1}{l}{Ours} \\ \midrule
\multicolumn{1}{c}{CIFAR-10}  & 
\multicolumn{1}{l}{}      & \multicolumn{1}{l}{}     \\ \midrule
VGG-8 W2A2                    & 92.09                     & \textbf{92.66}           \\
VGG-8 W4A4                    & 93.16                     & \textbf{93.66}           \\
ResNet-20 W2A2                & 91.80                     & \textbf{92.34}           \\
ResNet-20 W4A4                & 93.65                     & \textbf{94.3}            \\ \midrule
\multicolumn{1}{c}{CIFAR-100} & \multicolumn{1}{l}{}      & \multicolumn{1}{l}{}     \\ \midrule
ResNet-32 W2A2                & \textbf{70.62}            & 70.2                     \\
ResNet-32 W3A3                & 73.7                      & \textbf{74.09}           \\
ResNet-32 W4A4                & 74.76                     & \textbf{75.11}           \\
VGG-13 W2A2                   & 74.67                     & \textbf{76.59}           \\
VGG-13 W3A3                   & 75.27                     & \textbf{77.40}           \\
VGG-13 W4A4                   & 75.62                     & \textbf{77.97}           \\ \bottomrule
\end{tabular}
\end{table}

\begin{figure}[htb]

\centering
\includegraphics[width=.3\columnwidth]{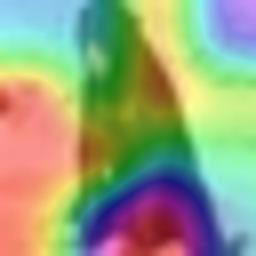}\hfill
\includegraphics[width=.3\columnwidth]{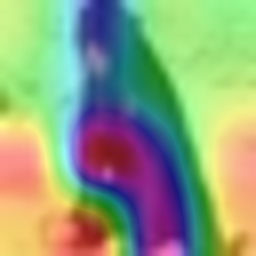}\hfill
\includegraphics[width=.3\columnwidth]{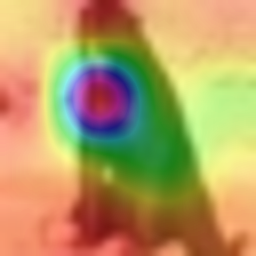}

\caption{The LayerCam \cite{layercam} heatmaps for the teacher network (left), the quantized model with data augmentations (center), and the quantized model without data augmentations (right). The color indicates the area of focus of the network, on a scale of red (low focus) to violet (high focus). }
\label{fig:heatmap}

\end{figure}

\boldhdr{Network Activations}
\cref{fig:heatmap} shows the LayerCam \cite{layercam} heatmap superimposed on the image being classified. This visualization highlights areas of relative focus from red (low focus) to violet (high focus), using the activations in the convolutional filters of the network, with each point further scaled by its weight and gradient. The values across every convolutional filter are averaged, yielding a single color for each pixel. All three images are classified by models trained using the same datasets, but we can see that the model trained under our chosen DA (center) generates a heatmap that aligns much more closely with the teacher's (left). Although this is only a single example, this demonstrates that models trained to the same quantization level can exhibit vastly different behaviors depending on the DA they were trained under. 

\boldhdr{Loss Terms}
\cref{tab:ablate_alpha} shows our results with different hyperparameter settings for loss terms trained under minimal data augmentations. Our loss is a weighted sum between the cross entropy loss from labels ($\gamma$) and loss from KD ($\alpha$)--in this case, we focus on traditional KD for simplicity. The case of $\gamma = 1, \alpha=0$ corresponds to QAT only without KD. We find that this method is significantly worse than KD with QAT when $\alpha =2$, matching the findings of previous work. In addition, we can see that in all experiments the model benefits from having labels available during training ($\gamma \neq 0$), similar to our findings in \cref{sec:kd}. Because we find the best configuration is $\gamma=1,\alpha=2$, we use this $1:2$ ratio for all our models trained using traditional KD. 
\begin{table}[htb]
\centering
\caption{Top-1 test accuracy evaluated across hyperparameter settings for loss terms. $\alpha$ represents the weight applied to traditional KD loss and $\gamma$ indicates weight attached to cross entropy loss from labels. These experiments were performed on CIFAR-100 using VGG-13 at W2A2 quantization.}
\label{tab:ablate_alpha}
\begin{tabular}{@{}lrrr@{}}
\toprule
          & \multicolumn{1}{l}{$\alpha = 0$} & \multicolumn{1}{l}{$\alpha = 1$} & \multicolumn{1}{l}{$\alpha = 2$} \\ \midrule
$\gamma = 0$ & -                          & 73.27                      & \multicolumn{1}{r}{73.73}                      \\
$\gamma = 1$ & 73.68                      & 73.74                      & \multicolumn{1}{r}{\textbf{74.67}}             \\ \bottomrule
\end{tabular}
\end{table}

\boldhdr{Network Initialization}
\cref{tab:ablate-init} shows the effects of network initialization for our QAT training. Because our teacher and student models are identical aside from the quantization, we can initialize the student's weights to the weights found in the teacher. We find that this initialization improves the resulting model performance upwards of 5\% compared to random initialization, which is a unique benefit offered by the process of self-supervised distillation. 

\begin{table}[htb]
\caption{Top-1 accuracy for quantized models under different initialization schemes. ``Teacher Weights'' indicates the student is initialized to the teacher's parameters, ``Random Weights'' indicates standard randomized model initialization.}
\label{tab:ablate-init}
\begin{tabular}{@{}lrr@{}}
\toprule
               & \multicolumn{2}{c}{Initialization}                                       \\ \midrule
               & \multicolumn{1}{l}{Teacher Weights} & \multicolumn{1}{l}{Random Weights} \\ \midrule
VGG-13 W2A2    & \textbf{74.67}                      & \multicolumn{1}{r}{66.79}                              \\
ResNet-32 W2A2 & \textbf{70.62}                      & \multicolumn{1}{r}{65.43}                              \\ \bottomrule
\end{tabular}
\end{table}

\section{Conclusion}
\label{sec:conclusion}
In this paper, we propose a new metric for ranking data augmentations within the context of KD-assisted QAT. The metric is cheap to compute, and it can greatly reduce the search space for selecting a data augmentation. We establish the effectiveness of the metric by evaluating over several popular data augmentations, showing that our selected augmentation performs well, and that it consistently outperforms benchmarks set by existing works combining KD with QAT. We show the flexibility of the approach by evaluating across a range of model architectures, bit-widths, quantizers, and KD methods, demonstrating the utility of guided DA search for KD in QAT.

{\small
\bibliographystyle{ieeenat_fullname}
\bibliography{11_references}
}

\end{document}